%% file: psc.tex
\documentclass{article}

\usepackage{PRIMEarxiv}

\usepackage[utf8]{inputenc} 
\usepackage[T1]{fontenc}    
\usepackage[hidelinks]{hyperref}       
\usepackage{url}            
\usepackage{booktabs}       
\usepackage{amsfonts}       
\usepackage{nicefrac}       
\usepackage{microtype}      
\usepackage{lipsum}
\usepackage{graphicx}
\graphicspath{{figures/}}     
\usepackage{cite}
\usepackage{amsmath,amssymb,amsfonts}
\usepackage{graphicx}
\usepackage{textcomp}
\usepackage{xcolor}
\usepackage{url}
\usepackage{subcaption}
\usepackage{multirow}
\usepackage{placeins}
\usepackage{tikz}
\usetikzlibrary{arrows.meta,positioning,fit,calc,backgrounds}

\author{
  Ivan Ovinnikov\quad Pascal Sutter\quad Christian Gehring\quad Jordis Herrmann \\
  ANYbotics AG \\
  Zürich, Switzerland\\
}

\begin{document}
\title{Predictive Safety Curricula for Robust Legged Locomotion}
\maketitle
\vspace{-.6cm}
\begin{abstract}
Rare but consequential failures can persist in learned locomotion policies for legged robots even when average task performance is high, in part because standard curricula primarily adapt task difficulty rather than the distribution of safety-critical experience. We introduce Predictive Safety Curricula (PSC), a framework for allocating locomotion training experience using learned predictions of future safety cost. PSC trains a distributional safety critic from policy rollouts and uses its predictions to prioritize both terrain contexts and previously encountered randomized events. The resulting curriculum modifies the training distribution while leaving the task reward and policy-optimization loss unchanged.
We evaluate PSC in controlled rough-terrain locomotion and in production locomotion systems. PSC improves reliability relative to standard terrain progression, advantage-based replay, and learning-progress curricula, with the largest gains on difficult terrain and under degraded observations. The same allocation principle transfers to two production locomotion stacks. On ANYmal-D hardware, PSC reduces shank-collision incidence by $63\%$ relative to the learning-progress curriculum across three matched training seeds, with a reduction in every seed. On a production stair-climbing platform, PSC eliminates observed shank collisions in the evaluated hardware trials. These results show that learned predictions of future safety cost can provide an effective signal for allocating training experience toward rare failure modes and improving locomotion reliability.
\end{abstract}


\section{Introduction}

Learning-based control has substantially expanded the capabilities of legged robots on complex and heterogeneous terrain. Despite this progress, policies
with strong average performance can still exhibit rare but consequential
unsafe outcomes, including lower-leg collisions and loss of balance under difficult conditions. Standard locomotion curricula prioritize task difficulty, average performance,
or learning progress, and may therefore overlook conditions that appear
equally successful on average but differ substantially in tail safety risk.
This creates a gap between capability-driven
training allocation and the rare safety-critical outcomes that
limit reliable deployment.

Curriculum learning provides a natural mechanism for reallocating training experience. Performance-based terrain curricula adapt task difficulty as locomotion competence improves~\cite{lee2020learning,rudin2022learning}, while Prioritized Level Replay and learning-progress curricula use policy-dependent signals to focus sampling on tasks with high estimated learning potential~\cite{jiang2021prioritized,li2026scaling}. These mechanisms support capability acquisition but do not explicitly target
elevated safety risk. Estimating such risk is challenging: failures may be sparse within individual terrain--command contexts and change throughout training. As the number of terrain and randomization variables grows, maintaining reliable empirical safety estimates for each configuration becomes increasingly data inefficient.

\begin{figure*}[t]
    \centering

    \begin{subfigure}[t]{0.40\textwidth}
        \centering
        \resizebox{\linewidth}{!}{\input{figures/fig1}}
        \caption{\textbf{Predictive safety allocation for context} $z=(\tau,\ell,b)$.
        Rollout states and costs train a shared safety-return critic. Its upper-tail
        risk is averaged into an episode score $\bar s_e$, which is aggregated
        by context and attached to event payloads encountered in that
        episode.}
        \label{fig:overview}
    \end{subfigure}
    \hfill
    \begin{subfigure}[t]{0.58\textwidth}
        \centering
        \includegraphics[width=\linewidth]{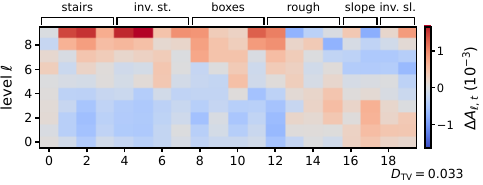}
        \caption{
        \textbf{Late-training context allocation under PSC relative to LP.}
        Mean difference in empirical terrain-context allocation,
        $\Delta A_{\ell,t}
        = A^{\mathrm{PSC}}_{\ell,t}
        - A^{\mathrm{LP}}_{\ell,t}$,
        averaged over $N=3$ training seeds.
        Positive values indicate contexts sampled more frequently by PSC.
        PSC induces a structured redistribution rather than uniformly
        favoring higher terrain levels.}
        \label{fig:allocation}
    \end{subfigure}

    \caption{Predictive safety allocation and its effect on the
    late-training context distribution.}
    \label{fig:overview_allocation}
\end{figure*}
We introduce Predictive Safety Curricula (PSC), a framework for directing locomotion training toward conditions associated with elevated predicted safety risk. PSC learns future safety cost from policy rollouts and uses these predictions to adapt subsequent training experience. Unlike safety critics used to constrain policy updates or intervene at deployment, PSC uses safety predictions to prioritize which experience is collected next: the task reward and loss function remain unchanged. The learned model supports both expected-cost and upper-tail prioritization and provides a common allocation signal across terrain and randomized environment variation.

We evaluate PSC in both controlled and production-level locomotion settings. On ANYmal-D, a quadrupedal robot developed by ANYbotics, PSC improves reliability relative to standard terrain progression, Prioritized Level Replay, and learning-progress curricula, with the largest gains under more demanding terrain and observation conditions. Controlled ablations examine the effect of experience allocation based on mean and upper-tail safety risk and compare predicted risk with direct empirical safety statistics, while critic diagnostics assess whether the learned model identifies future safety-relevant outcomes. We further evaluate PSC in a production stair-climbing stack on ANYmal-X, an explosion-safe quadruped where avoiding lower-leg impacts is particularly important for hardware integrity. There, PSC substantially reduces such contacts while preserving traversal performance.

Our contributions are threefold. \emph{First}, we introduce predictive safety allocation as a curriculum mechanism that uses a shared safety-return model to redistribute training experience, while the policy is still optimized for the original task reward using the unchanged reinforcement learning loss. \emph{Second}, we characterize the role of safety prediction and risk readout through controlled curriculum ablations and critic diagnostics. \emph{Third}, we validate the approach across a controlled locomotion benchmark, a production stair-climbing training stack, and repeated hardware trials with physically meaningful contact measurements.

\section{Method}
\label{sec:method}

PSC converts predictions of future safety exposure from completed rollouts
into priorities for subsequent training experience
(Fig.~\ref{fig:overview}). We represent the training distribution through
structured curriculum contexts and randomized environment events and use the
same learned safety signal to adapt both components. Context allocation can
operate alongside an existing capability curriculum that determines which
structured conditions are currently available, while event allocation adapts
the randomizations encountered within those conditions.

\subsection{Training Conditions and Safety Returns}
\label{sec:training_conditions}

We consider a structured curriculum context
$z=(\tau,\ell,b)\in\mathcal Z$, where $\tau$ denotes terrain type (e.g., stairs, slopes, or discrete obstacles),
$\ell\in\{1,\ldots,L\}$ the terrain difficulty level, and
$b\in\{1,\ldots,B\}$ an optional command bucket. Additional randomized
environment variables such as friction, joint positions, actuator properties,
or external perturbations, are represented by an event configuration $\xi$.
At curriculum update $j$, the joint distribution of training conditions is
factorized as
\begin{equation}
d_j(z,\xi)
=
p(\tau)\,
p_j^{\mathrm{ctx}}(\ell,b\mid\tau)\,
p_j^{\mathrm{evt}}(\xi\mid \tau,\ell,b),
\label{eq:training-distribution}
\end{equation}
where $p(\tau)$ denotes the fixed terrain-type assignment,
$p_j^{\mathrm{ctx}}$ controls allocation over level--command contexts within
each terrain type, and $p_j^{\mathrm{evt}}$ controls replayable randomized
events within the selected context. Environment variables not controlled by
either PSC sampler retain their original sampling procedures. This
factorization separates allocation across structured task contexts from
allocation over randomized environment events while allowing both to be
adapted using the same learned safety signal.
The locomotion policy $\pi_\theta(a_t\mid o_t)$ maps observation $o_t$ to
action $a_t$ and is trained using the original task reward. PSC introduces a
separate nonnegative safety cost
\begin{equation}
c_t
=
\sum_{m=1}^{M} w_m c_t^{(m)},
\qquad
w_m\geq0,
\qquad
G_t^c
=
\sum_{k=t}^{T-1}\gamma_c^{k-t}c_k,
\label{eq:safety_cost}
\end{equation}
where $c_t^{(m)}$ denotes task-specific safety component $m$, $w_m$ its
weight, and $\gamma_c\in(0,1]$ the safety discount. The discounted return
$G_t^c$ represents cumulative future safety exposure from time $t$ and
provides the prediction target for the safety critic. The particular safety
components and their parameters are properties of the locomotion task rather
than of PSC and are specified in Sec.~\ref{sec:experiments}.
PSC modifies the training-condition distribution $d_j$, thereby changing
the induced trajectory distribution used for policy optimization.
The safety cost trains an auxiliary critic whose predictions define
curriculum priorities; it is not added to the task reward or PPO policy
objective.

\subsection{Shared Distributional Safety Critic}
\label{sec:safety-critic}

Given rollout features $x_t$, a critic parameterized by $\psi$ predicts
$K$ conditional quantiles of the discounted safety return
$G_t^c$~\cite{dabney2018distributional}:
\begin{equation}
q_{\psi,i}(x_t)
\approx
F^{-1}_{G_t^c\mid x_t}(u_i),
\qquad
u_i=\frac{i-\tfrac12}{K},
\quad i=1,\ldots,K.
\label{eq:quantile-critic}
\end{equation}
Here $u_i$ is the probability level of the $i$-th predicted quantile.
Evenly spaced levels provide a uniform discretization of the conditional
safety-return distribution, allowing both mean and upper-tail statistics to
be obtained from the same critic. The function
$F^{-1}_{G_t^c\mid x_t}$ denotes the conditional quantile function, and
$x_t$ contains the state or observation features available to the safety
critic.

A single critic is trained from trajectories collected across the full
training distribution $d_j$. Safety observations from different curriculum
contexts and randomized environment configurations can therefore contribute
to a common predictive model when they induce similar risk-relevant states.
Temporal bootstrapping further propagates observed safety outcomes to states
preceding the corresponding interaction.
For an $n$-step segment ending at or before episode termination, the
distributional bootstrap targets are
\begin{equation}
y_{t,k}
=
\sum_{r=0}^{n-1}\gamma_c^r c_{t+r}
+
\gamma_c^n\chi_{t+n}q_{\bar\psi,k}(x_{t+n}),
\label{eq:quantile-target}
\end{equation}
where $\chi_{t+n}$ is $0$ at termination and $1$ otherwise, and
$\bar\psi$ denotes the Polyak-averaged target network parameters. The critic minimizes
the standard quantile-Huber regression loss~\cite{dabney2018distributional},
\begin{equation}
\mathcal L(\psi)
=
\mathbb E_t\!\left[
\frac{1}{K^2}
\sum_{i=1}^{K}\sum_{k=1}^{K}
\rho_{u_i}^{\kappa}
\!\left(y_{t,k}-q_{\psi,i}(x_t)\right)
\right],
\label{eq:quantile-loss}
\end{equation}
with Huber threshold $\kappa$.

Modeling the conditional safety-return distribution allows the same learned
predictor to support alternative risk summaries. We use an upper-tail
readout for PSC and evaluate the corresponding mean readout as an ablation in
Sec.~\ref{sec:safety_ablations}.


\subsection{From Safety Prediction to Curriculum Priority}
\label{sec:risk-readout}

Let $\alpha\in(0,1)$ denote the target upper-tail mass. The
state-conditioned upper-tail conditional value-at-risk (CVaR) is
\begin{equation}
R_\alpha(x_t)
=
\frac{1}{\alpha}
\int_{1-\alpha}^{1}
F^{-1}_{G_t^c\mid x_t}(u)\,\mathrm{d}u.
\label{eq:continuous-cvar}
\end{equation}
For the discrete quantile representation, let
$q_{\psi,(1)}(x)\leq\cdots\leq q_{\psi,(K)}(x)$ denote the predicted
quantile values ordered by magnitude and define
$k_\alpha=\lceil\alpha K\rceil$. We use
\begin{equation}
\begin{aligned}
\widehat R_\alpha^{\mathrm{disc}}(x)
&=
\frac{1}{k_\alpha}
\sum_{i=K-k_\alpha+1}^{K}
q_{\psi,(i)}(x),\\
\widehat\mu(x)
&=
\frac{1}{K}
\sum_{i=1}^{K}
q_{\psi,(i)}(x).
\end{aligned}
\label{eq:risk-readouts}
\end{equation}
We refer to $\widehat R_\alpha^{\mathrm{disc}}$ and $\widehat\mu$ as the
upper-tail and mean \emph{readouts} of the predicted safety-return
distribution. The safety cost determines which interactions contribute to
future safety exposure, while the readout determines how the predicted
distribution of that exposure enters curriculum allocation.

For each completed episode $e$ of length $T_e$, the selected state-level
readout $s$ is averaged along the trajectory:
\begin{equation}
\bar s_e
=
\frac{1}{T_e}
\sum_{t=0}^{T_e-1}
s(x_{e,t}),
\qquad
S_j(z)
=
\frac{1}{|\mathcal B_j(z)|}
\sum_{e\in\mathcal B_j(z)}
\bar s_e.
\label{eq:task-score}
\end{equation}
Here $\mathcal B_j(z)$ contains the recent scored episodes associated with
context $z$ at update $j$. PSC uses
$s=\widehat R_\alpha^{\mathrm{disc}}$ for upper-tail prioritization; the
mean-readout ablation uses $s=\widehat\mu$. The episode score $\bar s_e$
provides the common signal for the two allocation pathways. For context
allocation, recent episode scores are aggregated into $S_j(z)$; for event
allocation, the episode score remains associated with randomized event
realizations encountered during that trajectory.

As a non-predictive comparison, we additionally consider empirical CVaR
computed directly from realized episode safety returns. For a context
containing $n_z$ buffered episodes, the empirical priority averages the
largest $\max(1,\lceil\alpha n_z\rceil)$ values of $G_0^c$.
Sec.~\ref{sec:safety_ablations} compares this direct empirical statistic with
the learned state-conditioned priority.

\subsection{Safety-Prioritized Context Allocation}
\label{sec:safety-prioritized-curriculum}

For terrain type $\tau$, let $\mathcal E_{\tau,j}$ denote the set of
level--command pairs admitted by the underlying capability curriculum at
update $j$. PSC reallocates experience within this support by combining
uniform exploration with safety-prioritized sampling:
\begin{equation}
p_j^{\mathrm{ctx}}(\ell,b\mid\tau)
=
\varepsilon U_{\tau,j}(\ell,b)
+
(1-\varepsilon)
\frac{
\mathbf{1}\{(\ell,b)\in\mathcal E_{\tau,j}\}
P_j(\tau,\ell,b)
}{
\sum_{(\ell',b')\in\mathcal E_{\tau,j}}
P_j(\tau,\ell',b')
}.
\label{eq:safety-sampling}
\end{equation}
Here $U_{\tau,j}$ is uniform over the admissible set and $\varepsilon$
controls uniform exploration. The underlying capability curriculum therefore
determines which contexts are available, while PSC determines how experience
is distributed among them.

To maintain exploration, PSC combines the context score with terrain-level
staleness and coverage:
\begin{equation}
P_j(\tau,\ell,b)
=
\lambda_S\,\widetilde S_j(\tau,\ell,b)
+
\lambda_a\,\widetilde a_{\ell,j}
+
\lambda_u\,\widetilde u_{\ell,j}.
\label{eq:minmax-norm}
\end{equation}
Here, $a_{\ell,j}$ counts environment steps since level $\ell$ was last
selected, and $u_{\ell,j}=(1+n_{\ell,j})^{-1}$, where $n_{\ell,j}$ is its
cumulative scored-episode count. Tildes denote independent normalization
over the qualified support. Uninformative priorities yield uniform sampling
over the admissible support. The weights $\lambda_S,\lambda_a,\lambda_u$
and external capability curricula are specified in
Sec.~\ref{sec:experiments}.

\subsection{Safety-Prioritized Event Replay} 
\label{sec:event-replay} 
PSC adapts $p_j^{\mathrm{evt}}(\xi\mid z)$ by associating each sampled event realization $\xi$ with the safety score $\bar s_e$ of the episode in which it occurred. For each context, previously sampled event realizations are retained in a small replay buffer and preferentially resampled according to their scores. At each event invocation, PSC either reuses a context-matched realization or draws a new one, preserving exploration and providing a fallback when no replay samples are available. Context allocation and event replay therefore act on complementary factors of Eq.~\eqref{eq:training-distribution} using the same learned safety signal. 

\section{Experiments}
\label{sec:experiments}

We evaluate PSC across three increasingly realistic settings. The public Isaac Lab ANYmal-D benchmark provides a controlled environment in which policy architecture, reward design, optimization budget, and terrain progression are matched across methods; we use it for the main multi-seed comparison and mechanism ablations. All policies are trained with Proximal Policy Optimization (PPO)~\cite{schulman2017ppo}, with the policy-learning algorithm and objective held fixed across curriculum variants. We then assess PSC in the more complex production ANYmal-X stair-climbing stack, and finally evaluate trained production policies on hardware using physically meaningful contact outcomes.

\paragraph{Task and baselines.}
The ANYmal-D policy tracks commanded planar velocities and yaw rates over
procedurally generated rough terrain, with terrain difficulty represented by a
discrete level $\ell$. The underlying terrain curriculum determines which levels
are available for training, while the adaptive methods redistribute experience
within this support. For all adaptive methods, prioritized allocation is enabled
only after the terrain curriculum reaches its maximum level.

For PSC, we define an auxiliary safety signal that captures three common indicators of unsafe locomotion: excessive torso tilt, configurations in which the legs fold close to the robot base, and undesired thigh or shank contacts with the environment.
\begin{equation}
c_t
=
c_t^{\mathrm{ori}}
+
c_t^{\mathrm{dist}}
+
c_t^{\mathrm{und}}.
\end{equation}
The orientation term is
$c_t^{\mathrm{ori}}
=
[\cos(60^\circ)+g_{t,z}^{B}]_+$,
where $g_{t,z}^{B}$ is the vertical component of the unit gravity vector in the base frame.
The proximity term is
$c_t^{\mathrm{dist}}
=
[(d_{\mathrm{nom}}-d_t)/d_{\mathrm{nom}}]_+$,
with $d_{\mathrm{nom}}=0.35\,\mathrm{m}$ and $d_t$ the minimum distance from
the base-link origin to the hip, thigh, or shank link origins.
Finally,
$c_t^{\mathrm{und}}=\sum_{r\in\mathcal{B}_{\mathrm{und}}}
\mathbf{1}
\left\{
\max_h \lVert F_{t,h,r}\rVert_2 > 1\,\mathrm{N}
\right\},
$
where $\mathcal{B}_{\mathrm{und}}$ contains the thigh and shank bodies and $F_{t,h,r}$ denotes the contact-force
vector acting on body $r$ at contact-history sample $h$ of control step $t$.
All terms have unit weight and $\gamma_c=0.99$. The auxiliary safety cost itself
does not enter the task reward or PPO loss; its constituent failure and collision events are already included as penalties in the locomotion reward. 

We compare four curriculum strategies under the same locomotion learner.
\emph{Baseline} uses standard terrain progression without additional
prioritization. \emph{PLR} uses mean absolute generalized advantage and
combines score- and staleness-based priorities with $\rho=0.1$
~\cite{jiang2021prioritized}. Learning progress (\emph{LP}) uses the signed
change between consecutive episodic-return estimates followed by softmax
prioritization~\cite{li2026scaling}. \emph{PSC} derives priorities from
predicted safety return. To compare prioritization strategies, PLR, LP, and PSC share
the same admissible terrain--command support and randomized-event replay
pathway while retaining their respective priority definitions and sampling
rules. For PSC, we use $\lambda_S=0.7$, $\lambda_a=0.2$, and
$\lambda_u=0.1$ for predicted safety, staleness, and coverage.
We use a buffer size of 64 per replayable event with a replay probability of 0.4.

\paragraph{Safety-return estimator.}
The safety critic predicts $K=32$ quantiles of discounted safety return with
upper-tail mass $\alpha=0.05$. Following Sec.~\ref{sec:risk-readout}, the
upper-tail readout averages the largest
$k_\alpha=\lceil\alpha K\rceil=2$ quantiles, corresponding to $6.25\%$ of
the discrete representation. Episode and context scores follow
Eq.~\eqref{eq:task-score}.

The critic receives the same uncorrupted privileged observations as the PPO
value critic and is an independent ELU MLP with hidden widths
$(512,256,128)$. It uses one-step targets ($n=1$), $\gamma_c=0.99$,
quantile-Huber threshold $\kappa=1$, and Polyak coefficient $0.01$.
Each terrain-level--command context retains the latest $N=128$ episode
scores and enables prioritization after the first completed episode.
Predicted safety, staleness, and coverage are independently min--max
normalized, combined as above, and mixed with uniform sampling with
probability $0.3$.

\paragraph{Training and cohort provenance.}
All ANYmal-D methods use the same interaction and optimization budget:
8192 parallel environments, 24-step rollouts, and 3000 PPO updates, totaling
$589{,}824{,}000$ transitions. Each rollout is optimized for five epochs over
four minibatches using Adam with initial learning rate $10^{-3}$ and an
adaptive KL-based schedule.
For PSC, the safety critic is optimized jointly with the policy using the same
optimizer and schedule, updated once per PPO minibatch, with no separate
warm-up. Experiments use Isaac Lab \cite{mittal2025isaaclab}, and RSL-RL \cite{schwarke2025rslrl}. Unless
otherwise stated, results use six training seeds
$\{13,19,23,31,47,61\}$.

\paragraph{Evaluation protocol.}
We evaluate the final checkpoint after 3000 PPO updates using deterministic
actions. Each policy--condition pair runs for 3000 control steps in 1000
parallel environments at $50\,\mathrm{Hz}$ with a $40\,\mathrm{s}$ episode
timeout. Environments are reset after termination, and statistics are
computed over all episodes completed during evaluation; episodes truncated
by the evaluation cutoff are omitted.
Terrain and command assignments are shared across methods. Terrain cells are
assigned deterministically by environment index on a $10\times20$ grid, and
commands are drawn from
$
(v_x,v_y,\omega_z)\in
\{(0,0,0),(0.5,0,0),(1,0,0),(0.5,0.5,0),(0.5,0,0.5)\}.
$

To vary occupancy of difficult terrain, we use
\begin{equation}
p_v(\ell)\propto(\ell+1)^{v-1},
\qquad v\in\{1,2,3\},
\end{equation}
where $v=1$ is uniform and larger $v$ increasingly concentrates mass on
higher terrain levels. Observation noise doubles the nominal
policy-observation noise ranges. A fixed evaluation seed of 42 shares
simulator randomization across methods within each condition.

An episode fails on illegal torso contact (net contact force $>1\,\mathrm{N}$);
timeout at $40\,\mathrm{s}$ counts as success, and episodes truncated by the
evaluation cutoff are excluded.

\subsection{ANYmal-D Rough-Terrain Locomotion}
\label{sec:anymal_results}

Table~\ref{tab:anymal_success} summarizes the main ANYmal-D comparison.
PSC achieves the highest mean success across all six clean and
observation-noise conditions. PSC achieves higher mean success than PLR across the
evaluation grid, while the stronger LP curriculum narrows the gap. Relative
to LP, PSC improves success by $0.16$, $1.38$, and $0.50$ percentage points
under clean $v=1$, $v=2$, and $v=3$, respectively. The largest separation
occurs at $v=2$, where the evaluation distribution places greater mass on
more difficult terrain while retaining sufficient occupancy across the
curriculum.

\begin{table}[t]
\centering
\caption{ANYmal-D success (\%): mean $\pm$ sample standard deviation over
six training seeds. $v$ controls terrain occupancy. Bold denotes the highest
mean.}
\label{tab:anymal_success}
\small
\setlength{\tabcolsep}{2.5pt}
\begin{tabular}{@{}clcccc@{}}
\toprule
& Occupancy & Baseline & PLR & LP & PSC \\
\midrule
\multirow{3}{*}{\rotatebox[origin=c]{90}{Clean}}
& $v1$ & $91.94{\pm}1.56$ & $93.31{\pm}3.13$ &
$94.40{\pm}0.95$ & $\mathbf{94.56{\pm}1.08}$ \\
& $v2$ & $91.20{\pm}1.50$ & $92.29{\pm}4.65$ &
$93.81{\pm}1.00$ & $\mathbf{95.19{\pm}1.77}$ \\
& $v3$ & $89.07{\pm}2.06$ & $90.76{\pm}4.51$ &
$92.81{\pm}0.99$ & $\mathbf{93.31{\pm}1.36}$ \\
\midrule
\multirow{3}{*}{\rotatebox[origin=c]{90}{Noise}}
& $v1$ & $91.11{\pm}1.69$ & $91.44{\pm}2.98$ &
$92.27{\pm}1.78$ & $\mathbf{92.82{\pm}1.01}$ \\
& $v2$ & $88.81{\pm}2.65$ & $89.54{\pm}3.51$ &
$91.20{\pm}2.11$ & $\mathbf{92.53{\pm}1.74}$ \\
& $v3$ & $86.86{\pm}3.56$ & $87.78{\pm}4.36$ &
$89.08{\pm}2.30$ & $\mathbf{89.92{\pm}1.71}$ \\
\bottomrule
\end{tabular}
\vspace{-.5cm}
\end{table}

\paragraph{Reliability on difficult terrain.}
Expressed as failure rate, clean $v=2$ decreases from $6.19\%$ for LP to
$4.81\%$ for PSC, a $22.3\%$ relative reduction; at $v=3$, it decreases
from $7.19\%$ to $6.69\%$. Relative to PLR, failure decreases from
$7.71\%$ to $4.81\%$ at $v=2$ and from $9.24\%$ to $6.69\%$ at $v=3$.
Thus, PSC retains its advantage as evaluation shifts toward harder terrain,
with LP providing the stronger comparator.

\paragraph{Robustness to observation noise.}
PSC remains the highest-performing curriculum when policy observations are
corrupted by noise. Relative to LP, success improves by $0.55$, $1.33$, and $0.84$
percentage points under noisy $v=1$, $v=2$, and $v=3$, respectively. The
same ordering therefore persists under observation degradation, with the
largest difference again occurring at $v=2$. 

To examine how PSC changes the training distribution,
Fig.~\ref{fig:allocation} compares its late-training terrain-context
allocation with LP. PSC produces a structured, terrain-dependent
redistribution rather than uniformly favoring higher difficulty levels.
The seed-mean distributions differ by $0.033$ in terms of total variation distance, with
per-seed distances of $0.043$, $0.044$, and $0.045$.

\subsection{Safety Prediction and Curriculum Ablations}
\label{sec:safety_ablations}

We next examine how the safety signal used by PSC influences curriculum performance and whether the learned critic captures safety-relevant trajectory structure. All variants share the policy architecture, reward, sampling machinery, and training budget, with safety-cost changes stated explicitly where ablated. Curriculum results use the same six-seed cohort $\{13,19,23,31,47,61\}$ and six-condition ANYmal-D evaluation suite as the main comparison.

\paragraph{Risk readout and predictive estimation.}
We compare PSC (\emph{QC-CVaR}) with predictive mean and empirical tail
priorities. \emph{QC-mean} uses the same quantile critic but averages its
predicted quantiles, isolating the risk readout. \emph{Empirical CVaR} instead
computes an episode-level tail statistic from recent realized safety returns,
replacing the state-conditioned predictive score and its temporal aggregation
with a direct empirical priority.

Table~\ref{tab:safety_ablation} reports success averaged across evaluation
conditions within each seed. QC-CVaR achieves $93.05\pm1.14\%$, exceeding
QC-mean ($90.40\pm2.54\%$) and empirical CVaR ($91.38\pm3.17\%$) by paired
differences of $2.66\pm2.96$ and $1.68\pm3.80$ percentage points,
respectively. QC-CVaR is higher than QC-mean on five of six seeds, while the
empirical comparison is less consistent. The results favor predictive
upper-tail prioritization in this cohort, although seed-level variability
precludes a strong causal attribution to the readout alone. QC-mean is also
sensitive to the safety-cost composition: removing the undesired-contact term
increases success to $91.17\pm1.11\%$.

The ablations support predictive safety as a useful allocation signal without
isolating a unique contribution from the upper-tail readout. Empirical tail statistics yield lower mean success and greater across-seed
variability than learned state-conditioned priorities in this cohort.

\paragraph{Predictive quality of the safety critic.}
We evaluate the critic independently of final policy performance. Return
prediction is measured by Spearman correlation between the mean prediction
$\widehat{\mu}(x_t)$ and Monte-Carlo discounted safety return $G_t^c$.
To test whether high predicted risk identifies states preceding unsafe events,
we use the state-level CVaR score to rank whether an event occurs within
the next $H=100$ control steps ($2\,\mathrm{s}$). We consider two outcomes: a \emph{collision}, defined as
the terminating illegal base contact (net force $>1\,\mathrm{N}$), and a
\emph{contact}, defined as any thigh or shank contact above $1\,\mathrm{N}$.
Termination or timeout truncates the horizon, and incomplete windows at the
evaluation cutoff are omitted. We report the area under the precision--recall curve (AUPRC) minus event
prevalence and top-decile lift, defined as the event rate among the
highest-scoring $10\%$ of valid states divided by the overall event rate.

\begin{table*}[t]
    \centering

    \begin{minipage}[t]{0.42\textwidth}
        \vspace{0pt}
        \centering
        \captionof{table}{
            ANYmal-D safety-priority ablations. Success is averaged over the six
            evaluation conditions within each seed and reported as mean $\pm$
            sample standard deviation over six training seeds. $\Delta$ denotes
            the paired difference relative to QC-CVaR.
        }
        \label{tab:safety_ablation}

        \small
        \setlength{\tabcolsep}{4pt}
        \begin{tabular}{@{}lcc@{}}
            \toprule
            Priority
            & Success (\%) $\uparrow$
            & $\Delta$ (pp) \\
            \midrule
            QC-mean
            & $90.40 \pm 2.54$
            & $-2.66 \pm 2.96$ \\
            Empirical CVaR
            & $91.38 \pm 3.17$
            & $-1.68 \pm 3.80$ \\
            QC-CVaR (PSC)
            & $\mathbf{93.05 \pm 1.14}$
            & --- \\
            \bottomrule
        \end{tabular}
    \end{minipage}
    \hfill
    \begin{minipage}[t]{0.55\textwidth}
        \vspace{0pt}
        \centering
        \captionof{table}{
            Safety-critic diagnostics over six seeds: mean $\pm$ sample standard
            deviation and $95\%$ confidence interval. AUPRC gain is measured
            relative to event prevalence; top-decile lift has reference value one.
        }
        \label{tab:critic_diagnostics}

        \small
        \setlength{\tabcolsep}{3pt}
        \begin{tabular}{@{}lcc@{}}
            \toprule
            Metric & Mean $\pm$ s.d. & $95\%$ CI \\
            \midrule
            Spearman$(\widehat{\mu},G_t^c)$
            & $0.773 \pm 0.018$
            & $[0.754,0.792]$ \\
            Collision AUPRC gain
            & $0.079 \pm 0.015$
            & $[0.063,0.095]$ \\
            Collision top-decile lift
            & $6.88 \pm 0.19$
            & $[6.68,7.08]$ \\
            Contact AUPRC gain
            & $0.461 \pm 0.009$
            & $[0.452,0.470]$ \\
            Contact top-decile lift
            & $2.81 \pm 0.09$
            & $[2.72,2.90]$ \\
            \bottomrule
        \end{tabular}
    \end{minipage}

    \vspace{-0.3cm}
\end{table*}

As shown in Table~\ref{tab:critic_diagnostics}, 
the mean prediction correlates strongly with realized discounted safety return
(Spearman $\rho=0.773$). High CVaR scores also concentrate subsequent unsafe
events: states in the highest-scoring decile have $6.88\times$ the overall
collision rate and $2.81\times$ the overall thigh-or-shank contact rate.
AUPRC exceeds event prevalence by $0.079$ for collisions and $0.461$ for
contacts. These results show that the critic ranks visited states by future
safety exposure, supporting its use as a curriculum-priority signal.
\paragraph{Tail-mass sensitivity.} We further examine sensitivity to the upper-tail mass $\alpha$. Relative to the default configuration ($\alpha=0.05$, $93.05 \pm 1.14\%$), setting $\alpha=0.01$, $0.10$, and $0.20$ yields $92.25{\pm}2.88\%$, $91.80{\pm}2.06\%$, and $91.28{\pm}1.31\%$ mean success, respectively. Performance therefore decreases as the readout is broadened beyond the default tail mass, while the more concentrated $\alpha=0.01$ variant remains closer to PSC. Among the tested settings, $\alpha=0.05$ achieves the highest mean reliability.

\subsection{ANYmal-X Stair Climbing} 
\label{sec:anymal_x_stairs} 

We next evaluate PSC in the production stair-climbing training stack used for
ANYmal-X. Compared with ANYmal-D, this setting uses a different embodiment and
a distinct production locomotion stack, providing a complementary test of
predictive safety allocation under a substantially different training system.

\paragraph{Experimental setup.}
We compare PSC with the baseline curriculum, PLR, and LP using the same
stair-climbing learner and matched training budget. Teacher policies are
evaluated on a held-out stair-climbing task, distinct from the training task.
Each episode consists of a single staircase traversal toward a commanded
planar goal; timeout, out-of-bounds motion, illegal base contact, and excessive
body tilt are treated as failures.
We evaluate six settings
$\{\texttt{v0},\texttt{v1},\texttt{v1.5},\texttt{v2},
\texttt{v2.5},\texttt{v3}\}$ that jointly increase commanded speed and stair
difficulty. Evaluation conditions are matched across methods. Each method is
trained with three seeds, and results are reported as mean $\pm$ sample
standard deviation over training seeds.

\paragraph{Student distillation.}
The production policies are deployed after teacher--student distillation, and
the PSC advantage is largely preserved through this transfer. Across the six
stair-climbing settings, PSC students achieve
$87.76{\pm}1.70\%$ mean success, compared with
$86.72{\pm}0.71\%$ for the baseline and $87.06{\pm}1.06\%$ for PLR.
A student distilled from the PSC teacher without PSC-based prioritization
during distillation reaches $87.52{\pm}1.39\%$, indicating that most of the
reliability improvement is captured by the teacher and retained through the
production distillation pipeline.

\begin{table}[t]
    \centering
    \caption{ANYmal-X stair-climbing teacher success (\%): mean $\pm$
    sample standard deviation over three training seeds. The final row
    averages settings within each seed.}
    \label{tab:anymal_x_stairs}
    \small
    \setlength{\tabcolsep}{2.5pt}
    \begin{tabular}{@{}lcccc@{}}
        \toprule
        Setting & Baseline & PLR & LP & PSC \\
        \midrule
        \texttt{v0}
        & $99.99 \pm 0.01$ & $99.98 \pm 0.02$
        & $99.99 \pm 0.02$ & $\mathbf{100.00 \pm 0.00}$ \\
        \texttt{v1}
        & $99.65 \pm 0.17$ & $99.69 \pm 0.07$
        & $99.67 \pm 0.19$ & $\mathbf{99.85 \pm 0.03}$ \\
        \texttt{v1.5}
        & $96.78 \pm 1.35$ & $96.32 \pm 0.74$
        & $96.06 \pm 0.60$ & $\mathbf{97.35 \pm 0.87}$ \\
        \texttt{v2}
        & $89.96 \pm 1.76$ & $90.13 \pm 1.57$
        & $90.54 \pm 0.30$ & $\mathbf{91.90 \pm 0.86}$ \\
        \texttt{v2.5}
        & $76.14 \pm 4.00$ & $76.29 \pm 3.03$
        & $76.14 \pm 1.49$ & $\mathbf{79.16 \pm 4.03}$ \\
        \texttt{v3}
        & $57.75 \pm 3.06$ & $59.11 \pm 2.22$
        & $59.88 \pm 2.53$ & $\mathbf{62.46 \pm 3.74}$ \\
        \midrule
        Mean
        & $86.71 \pm 1.69$ & $86.92 \pm 1.24$
        & $87.04 \pm 0.60$ & $\mathbf{88.45 \pm 1.54}$ \\
        \bottomrule
    \end{tabular}
    \vspace{-.5cm}
\end{table}

\paragraph{Reliability in the production training stack.}
PSC achieves the highest mean success across all six stair-climbing
conditions (Table~\ref{tab:anymal_x_stairs}). Averaged across settings,
success reaches $88.45\%$, compared with $86.71\%$ for the baseline,
$86.92\%$ for PLR, and $87.04\%$ for LP. The corresponding improvements are
$1.74$, $1.53$, and $1.41$ percentage points, respectively.

The separation is larger in the more demanding settings. Relative
to LP, the strongest comparator on average, PSC improves success by
$1.36$ percentage points at \texttt{v2}, $3.02$ at \texttt{v2.5}, and
$2.58$ at \texttt{v3}. At the hardest setting, PSC succeeds in $62.46\%$ of
episodes, compared with $59.88\%$ for LP and $57.75\%$ for the baseline.
The same ordering observed in the controlled ANYmal-D benchmark is also observed in the production stair-climbing training stack, with the clearest
differences appearing in the more difficult evaluation regimes.

\FloatBarrier

\subsection{Hardware Evaluation}
\label{sec:hardware}

\begin{figure*}[t]
    \centering
    \includegraphics[width=\textwidth]{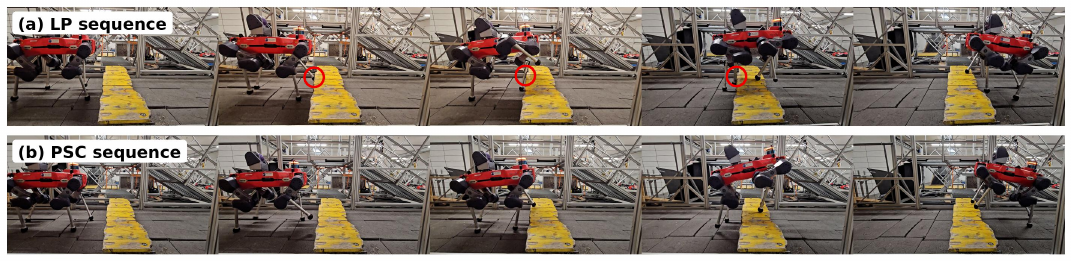}
    \caption{
    Hardware evaluation on ANYmal-D.
    Representative open-step traversals under
    (a) LP and (b) PSC, shown as five-frame sequences from matched-view
    recordings. Quantitative results are reported in
    Table~\ref{tab:hardware_amd}. Red circles mark annotated shank contacts.
    }
    \label{fig:hardware}
\end{figure*}

Simulation success does not fully characterize physical locomotion: policies
that complete the same task may differ substantially in how they interact
with the terrain. We therefore evaluate distilled production policies on
ANYmal-D open-step walking (Fig.~\ref{fig:hardware}) and on the production
stair-climbing platform, using lower-leg contact as the primary hardware
outcome. All PSC components other than the locomotion policy are discarded
after training.

\paragraph{Protocol and metrics.}
Deployed policies are obtained through the same matched teacher--student
pipeline. Contacts are manually annotated from recorded trials. A trial is
\emph{shank-positive} if at least one shank collision occurs; additional shank
contacts within the same traversal contribute to the total event count but not
to incidence. Foot scuffs are reported separately and are not classified as
collision failures in the simulation safety definition. Policies are evaluated
sequentially, with the robot software restarted when switching policies.

\paragraph{ANYmal-D open-step traversal.}
We evaluate production rough-terrain policies trained with the same
teacher--student pipeline for three curricula. For each method, three matched
seeds are evaluated on hardware using the final checkpoints, without
hardware-based selection or fine-tuning; LP and PSC differ from the baseline
only in their training-time prioritization. Policies traverse a $25\,\mathrm{cm}$-high, $40\,\mathrm{cm}$-deep open step at $(v_x,v_y,\omega_z)=(0.7,0,0)$, alternating direction between crossings. Each seed completes 100 one-way crossings, yielding 300 crossings per method. PSC reduces shank-positive crossings to $10.7\pm4.9$ per 100 crossings, compared with $25.7\pm9.1$ for the baseline and $29.0\pm5.3$ for LP (Table~\ref{tab:hardware_amd}), corresponding to a $63.2\%$ reduction relative to LP. PSC has fewer shank-positive crossings than both alternatives in all three seeds.
Foot scuffs increase relative to LP ($29.0\pm9.6$ vs. $14.3\pm1.5$ per 100 crossings), while remaining comparable to the baseline ($31.0\pm24.6$). Because foot contacts are excluded from $\mathcal{B}_{\mathrm{und}}$, this suggests that PSC shifts contact away from the explicitly penalized shank rather than uniformly suppressing lower-limb contact, highlighting the specificity of the safety-cost definition.

\paragraph{Production stair climbing.}
We additionally deploy distilled students from
Sec.~\ref{sec:anymal_x_stairs} on the production stair-climbing system.
Policies traverse a fixed staircase under matched conditions using the same
commanded speed for both policies. Contacts are annotated using the same
procedure as for ANYmal-D.
Across 50 ascent--descent pairs, the baseline policy produces shank collisions
in $33/50$ trials and 41 individual shank-collision events, whereas no shank
collision is observed with PSC. Foot scuffs decrease from 14 to 7, and
completion increases from $49/50$ to $50/50$; the single unsuccessful baseline
traversal ends with the robot stuck in the stairs. Observed mean traversal
speeds are effectively identical between the two policies.

\begin{table*}[t]
    \centering

    \begin{minipage}[t]{0.49\textwidth}
        \vspace{0pt}
        \centering
        \captionof{table}{
            ANYmal-D open-step hardware evaluation over three matched
            training seeds, with 100 one-way crossings per seed and method.
            Values are mean $\pm$ sample s.d. across three seeds.
        }
        \label{tab:hardware_amd}

        \small
        \setlength{\tabcolsep}{3pt}
        \begin{tabular}{@{}lccc@{}}
            \toprule
            Metric & Baseline & LP & PSC \\
            \midrule
            Shank-positive crossings
            & $25.7{\pm}9.1$
            & $29.0{\pm}5.3$
            & $\mathbf{10.7}{\pm}4.9$ \\
            Foot-scuff events
            & $31.0{\pm}24.6$
            & $\mathbf{14.3}{\pm}1.5$
            & $29.0{\pm}9.6$ \\
            \bottomrule
        \end{tabular}
    \end{minipage}
    \hfill
    \begin{minipage}[t]{0.48\textwidth}
        \vspace{0pt}
        \centering
        \captionof{table}{
            Production stair-climbing hardware evaluation over $50$
            ascent--descent pairs per policy. Shank-positive trials contain
            at least one lower-leg collision.
        }
        \label{tab:hardware_amx}

        \small
        \setlength{\tabcolsep}{4pt}
        \begin{tabular}{@{}lcc@{}}
            \toprule
            Metric & Baseline & PSC \\
            \midrule
            Shank-positive trials
            & $33/50$ & $\mathbf{0/50}$ \\
            Shank-collision events
            & $41$ & $\mathbf{0}$ \\
            Foot-scuff events
            & $14$ & $\mathbf{7}$ \\
            Completion
            & $49/50$ & $\mathbf{50/50}$ \\
            Stuck events
            & $1$ & $0$ \\
            \bottomrule
        \end{tabular}
    \end{minipage}

    \vspace{-0.3cm}
\end{table*}

\FloatBarrier

\section{Related Work}
\label{sec:related-work}

\paragraph{Curriculum learning and adaptive environment sampling.}
Curriculum learning adapts the training distribution according to competence,
learning progress, or task difficulty~\cite{narvekar2020curriculum,portelas2020automatic}.
Performance-based terrain progression is widely used in legged locomotion
~\cite{lee2020learning,rudin2022learning}, while perceptive methods additionally
vary terrain and sensing conditions~\cite{miki2022learning}. More general
adaptive sampling methods prioritize environments using learning progress,
policy uncertainty, or learning potential, including
ALP-GMM~\cite{portelas2020teacher}, PLR~\cite{jiang2021prioritized},
ACCEL~\cite{parkerholder2022evolving}, and Active Domain Randomization
~\cite{mehta2020active}. Recent locomotion-specific approaches include
LP-ACRL~\cite{li2026scaling} and HACL~\cite{mishra2025hacl}. ADD learns an
environment-conditioned task-return distribution and uses a CVaR-derived regret
signal to guide a diffusion-based environment generator~\cite{chung2024add}. PSC
instead uses predictions of future safety exposure to reallocate training
experience over structured contexts and randomized environment events.

\paragraph{Risk-aware curricula and failure-driven sampling.}
RACGEN concentrates an auxiliary sampler on low-return contexts in
heavy-tailed task distributions~\cite{koprulu2023riskaware}, while CeSoR
combines adverse-condition sampling with a soft-risk policy objective
~\cite{greenberg2022efficient}. Other approaches incorporate safety directly
into curriculum construction: curriculum induction learns when to deploy a
reset controller~\cite{turchetta2020safe}, while Safety-Prioritizing Curricula
initially favor tasks with fewer constraint violations~\cite{koprulu2025safety}.
Failure predictors have also been used to search for rare catastrophic
outcomes during evaluation~\cite{uesato2019rigorous}, and TACL learns
transition difficulty from success and failure labels to generate
capability-matched trajectories~\cite{liu2026trajectory}. PSC differs in
learning a shared distribution of future safety cost from rollout states and
using this prediction to determine where subsequent training experience is
allocated.

\paragraph{Safety critics and distributional risk.}
Safety critics are commonly used to modify policy behavior or constrain policy
optimization. CPO imposes cumulative-cost constraints on the policy
update~\cite{achiam2017constrained}; Recovery RL combines a safety critic with
a recovery policy~\cite{thananjeyan2021recovery}; and Agile But Safe uses a
reach-avoid value function for locomotion recovery~\cite{he2024agile}.
Distributional safety critics further apply risk measures such as CVaR to
reward--safety trade-offs~\cite{yang2022safety}. PSC builds on distributional
RL~\cite{bellemare2017distributional,dabney2018distributional,dabney2018implicit}
and the standard CVaR risk measure~\cite{rockafellar2002conditional}, but uses
the resulting safety prediction for a different intervention: selecting which
simulated experience is collected next. Unlike risk-sensitive locomotion
methods~\cite{schneider2024riskaware,shi2023robust}, the safety model does not
modify the task reward or PPO loss and is not used at deployment.

\section{Conclusion}
\label{sec:conclusion}

Closing the final gap toward near-perfect reliability is increasingly challenging, yet indispensable for the safe deployment of learned control policies on legged robots. Toward this goal, we introduced Predictive Safety Curricula (PSC), which uses predictions of future safety exposure to allocate locomotion training experience. A distributional safety critic drives both context allocation and randomized-event replay, concentrating experience in regions of the training distribution associated with elevated predicted risk. Across all evaluated ANYmal-D terrain and observation conditions, PSC achieves the highest mean reliability among the compared curricula. Ablations favor learned predictive priorities over direct empirical tail statistics, while critic diagnostics show that the learned scores meaningfully rank realized safety return and enrich subsequent collision and contact events.

The allocation principle transfers to production locomotion stacks and through teacher--student distillation. On ANYmal-D hardware, PSC reduces shank-collision incidence by $63\%$ relative to the learning-progress curriculum across three matched training seeds, with a reduction in every seed. On the production stair-climbing platform, no shank collision is observed with PSC in the evaluated hardware trials, compared with $33/50$ shank-positive trials for the baseline, at similar observed mean traversal speeds. Together, these results show that predictions of future safety exposure can provide a practical signal for deciding which experience a locomotion policy should train on.

\paragraph*{Limitations and outlook.}
PSC currently relies on a hand-specified safety cost and can adapt experience only through environment variables exposed to the curriculum. The hardware results illustrate the first limitation directly: reducing targeted shank contacts does not necessarily suppress other interactions, such as foot scuffs that are absent from the safety cost. Future work should therefore consider structured safety predictions that preserve multiple failure modes rather than collapsing safety exposure into a single fixed scalar objective. On the allocation side, a natural extension is to move beyond reweighting and replay within a predefined environment parameterization toward actively discovering or generating conditions predicted to expose policy weaknesses. Together, richer safety representations and active failure-condition discovery would extend PSC from prioritizing safety-relevant experience within an existing curriculum to shaping the training distribution itself around the remaining reliability gaps.

\section*{Acknowledgments}
This work was supported in part by EuroHPC under project
EHPC-AIF-2026FL01-031.
\newpage
\bibliographystyle{unsrt}  
\bibliography{refs}

\end{document}

%% file: figures/fig1.tex

\begin{tikzpicture}[
  font=\scriptsize,
  >={Latex[length=1.4mm,width=1.2mm]},
  blk/.style={draw,rounded corners=1.2pt,align=center,
              inner sep=2.2pt,minimum height=6.5mm},
  psc/.style={blk,fill=black!6},
  lnk/.style={->,thick,black!75},
  fb/.style={lnk,dashed},
  ann/.style={font=\scriptsize,align=center,inner sep=1.5pt}
]

\node[inner sep=0pt] (D) at (0.65,0)
  {$d_j(z,\xi)=$};

\node[blk,text width=1.15cm] (F) at (2.00,0)
  {Fixed\\$p(\tau)$};

\node[psc,text width=2.20cm] (CTX) at (4.30,0)
  {Contexts\\$p_j^{\mathrm{ctx}}(\ell,b\mid\tau)$};

\node[psc,text width=2.20cm] (EVT) at (6.95,0)
  {Events\\$p_j^{\mathrm{evt}}(\xi\mid z)$};

\node at (2.86,0) {$\times$};
\node at (5.63,0) {$\times$};

\begin{scope}[on background layer]
  \node[draw,black!40,rounded corners=2pt,inner sep=3pt,
        fit=(D)(F)(CTX)(EVT)] (DIST) {};
\end{scope}

\node[blk,text width=2.80cm] (R) at (4.30,-1.30)
  {Simulated rollouts};

\node[blk,text width=1.75cm] (PPO) at (1.30,-2.50)
  {PPO update};

\node[blk,text width=1.75cm] (POL) at (1.30,-3.60)
  {Policy $\pi_\theta$\\deployed alone};

\draw[lnk] (4.30,0 |- DIST.south) -- (R.north);

\draw[lnk] (R.west)
  -| node[pos=0.75,left,ann] {$r_t$} (PPO.north);

\draw[lnk] (PPO) -- (POL);

\draw[lnk] (POL.west)
  -- (0.20,-3.60)
  -- (0.20,-0.77)
  -| (R.north west);

\node[psc,text width=2.80cm] (Q) at (4.30,-2.50)
  {Distributional safety-return critic\\$q_{\psi,i}(x_t)$};

\node[psc,text width=2.80cm] (SCORE) at (4.30,-3.60)
  {Upper-tail risk\\
   $\widehat R_{\alpha}^{\mathrm{disc}}(x_t)\;\to\;\bar s_e$};

\draw[lnk] (R) -- node[right,ann] {$x_t,c_t$} (Q);
\draw[lnk] (Q) -- (SCORE);

\node[psc,text width=3.15cm,minimum height=10.5mm] (OUTER)
  at (2.20,-4.95)
  {\textbf{Context allocation}\\
   aggregate risk by terrain--command context};

\node[psc,text width=3.15cm,minimum height=10.5mm] (INNER)
  at (6.20,-4.95)
  {\textbf{Event replay}\\
   replay high-risk events in matched contexts};
\coordinate (SPLIT) at (4.30,-4.15);

\draw[thick,black!75] (SCORE.south) -- (SPLIT);
\fill[black!75] (SPLIT) circle (0.65pt);

\draw[lnk] (SPLIT) -| (OUTER.north);
\draw[lnk] (SPLIT) -| (INNER.north);

\draw[lnk] (R.east)
  -- node[below,ann] {encountered\\event payloads} (8.06,-1.30)
  |- (INNER.east);


\draw[fb] (OUTER.south)
  -- (2.20,-5.80)
  -- (-0.20,-5.80)
  -- (-0.20,0.85)
  -| (CTX.north);

\draw[fb] (INNER.south)
  -- (6.20,-5.80)
  -- (8.55,-5.80)
  -- (8.55,0.85)
  -| (EVT.north);

\end{tikzpicture}


%% file: refs.bib
@article{narvekar2020curriculum,
 author={Sanmit Narvekar and Bei Peng and Matteo Leonetti and Jivko Sinapov and Matthew E. Taylor and Peter Stone},
 title={Curriculum Learning for Reinforcement Learning Domains: A Framework and Survey},
 journal={Journal of Machine Learning Research}, volume={21}, number={181}, pages={1--50}, year={2020}}

@article{portelas2020automatic,
 author={R{\'e}my Portelas and C{\'e}dric Colas and Lilian Weng and Katja Hofmann and Pierre-Yves Oudeyer},
 title={Automatic Curriculum Learning for Deep {RL}: A Short Survey},
 journal={arXiv preprint arXiv:2003.04664}, year={2020}}

@article{lee2020learning,
 author={Joonho Lee and Jemin Hwangbo and Lorenz Wellhausen and Vladlen Koltun and Marco Hutter},
 title={Learning Quadrupedal Locomotion over Challenging Terrain},
 journal={Science Robotics}, volume={5}, number={47}, pages={eabc5986}, year={2020}}

@inproceedings{rudin2022learning,
 author={Nikita Rudin and David Hoeller and Philipp Reist and Marco Hutter},
 title={Learning to Walk in Minutes Using Massively Parallel Deep Reinforcement Learning},
 booktitle={Conference on Robot Learning}, series={Proceedings of Machine Learning Research}, volume={164}, pages={91--100}, year={2022}}

@article{miki2022learning,
 author={Takahiro Miki and Joonho Lee and Jemin Hwangbo and Lorenz Wellhausen and Vladlen Koltun and Marco Hutter},
 title={Learning Robust Perceptive Locomotion for Quadrupedal Robots in the Wild},
 journal={Science Robotics}, volume={7}, number={62}, pages={eabk2822}, year={2022}}

@article{li2026scaling,
  title   = {Scaling Rough Terrain Locomotion With Automatic Curriculum Reinforcement Learning},
  author  = {Li, Ziming and Li, Chenhao and Hutter, Marco},
  journal = {IEEE Robotics and Automation Letters},
  volume  = {11},
  number  = {8},
  pages   = {9295--9302},
  year    = {2026},
  doi     = {10.1109/LRA.2026.3703486}
}

@article{mishra2025hacl,
 author={Prakhar Mishra and Amir Hossain Raj and Xuesu Xiao and Dinesh Manocha},
 title={{HACL}: History-Aware Curriculum Learning for Fast Locomotion},
 journal={arXiv preprint arXiv:2505.18429}, year={2025}}

@inproceedings{portelas2020teacher,
 author={R{\'e}my Portelas and C{\'e}dric Colas and Katja Hofmann and Pierre-Yves Oudeyer},
 title={Teacher Algorithms for Curriculum Learning of Deep {RL} in Continuously Parameterized Environments},
 booktitle={Conference on Robot Learning}, series={Proceedings of Machine Learning Research}, volume={100}, pages={835--853}, year={2020}}

@inproceedings{jiang2021prioritized,
 author={Minqi Jiang and Edward Grefenstette and Tim Rockt{\"a}schel},
 title={Prioritized Level Replay},
 booktitle={International Conference on Machine Learning}, series={Proceedings of Machine Learning Research}, volume={139}, pages={4940--4950}, year={2021}}

@inproceedings{parkerholder2022evolving,
 author={Jack Parker-Holder and Minqi Jiang and Michael Dennis and Mikayel Samvelyan and Jakob Foerster and Edward Grefenstette and Tim Rockt{\"a}schel},
 title={Evolving Curricula with Regret-Based Environment Design},
 booktitle={International Conference on Machine Learning}, series={Proceedings of Machine Learning Research}, volume={162}, pages={17473--17498}, year={2022}}

@inproceedings{mehta2020active,
 author={Bhairav Mehta and Manfred Diaz and Florian Golemo and Christopher J. Pal and Liam Paull},
 title={Active Domain Randomization},
 booktitle={Conference on Robot Learning}, series={Proceedings of Machine Learning Research}, volume={100}, pages={1162--1176}, year={2020}}

@inproceedings{koprulu2023riskaware,
 author={Cevahir Koprulu and Thiago D. Sim{\~a}o and Nils Jansen and Ufuk Topcu},
 title={Risk-Aware Curriculum Generation for Heavy-Tailed Task Distributions},
 booktitle={Conference on Uncertainty in Artificial Intelligence}, series={Proceedings of Machine Learning Research}, volume={216}, pages={1132--1142}, year={2023}}

@inproceedings{greenberg2022efficient,
 author={Ido Greenberg and Yinlam Chow and Mohammad Ghavamzadeh and Shie Mannor},
 title={Efficient Risk-Averse Reinforcement Learning},
 booktitle={Advances in Neural Information Processing Systems}, year={2022}}

@inproceedings{turchetta2020safe,
 author={Matteo Turchetta and Andrey Kolobov and Shital Shah and Andreas Krause and Alekh Agarwal},
 title={Safe Reinforcement Learning via Curriculum Induction},
 booktitle={Advances in Neural Information Processing Systems}, year={2020}}

@inproceedings{koprulu2025safety,
 author={Cevahir Koprulu and Thiago D. Sim{\~a}o and Nils Jansen and Ufuk Topcu},
 title={Safety-Prioritizing Curricula for Constrained Reinforcement Learning},
 booktitle={International Conference on Learning Representations}, year={2025}}

@inproceedings{uesato2019rigorous,
 author={Jonathan Uesato and Ananya Kumar and Csaba Szepesv{\'a}ri and Tom Erez and Avraham Ruderman and Keith Anderson and Krishnamurthy Dvijotham and Nicolas Heess and Pushmeet Kohli},
 title={Rigorous Agent Evaluation: An Adversarial Approach to Uncover Catastrophic Failures},
 booktitle={International Conference on Learning Representations}, year={2019}}

@article{liu2026trajectory,
 author={Rocky Liu and Tengyu Liu and Baoxiong Jia and Fangwei Zhong and Xinyi Tong and Hongzhao Xie and Siyuan Huang},
 title={Trajectory-Level Automatic Curriculum Learning for Legged Locomotion on Unstructured Terrain},
 journal={arXiv preprint arXiv:2608.16164}, year={2026}}

@inproceedings{achiam2017constrained,
 author={Joshua Achiam and David Held and Aviv Tamar and Pieter Abbeel},
 title={Constrained Policy Optimization},
 booktitle={International Conference on Machine Learning}, series={Proceedings of Machine Learning Research}, volume={70}, pages={22--31}, year={2017}}

@article{thananjeyan2021recovery,
 author={Brijen Thananjeyan and Ashwin Balakrishna and Suraj Nair and Michael Luo and Krishnan Srinivasan and Minho Hwang and Joseph E. Gonzalez and Julian Ibarz and Chelsea Finn and Ken Goldberg},
 title={Recovery {RL}: Safe Reinforcement Learning with Learned Recovery Zones},
 journal={IEEE Robotics and Automation Letters}, year={2021}}

@inproceedings{he2024agile,
 author={Tairan He and Chong Zhang and Wenli Xiao and Guanqi He and Changliu Liu and Guanya Shi},
 title={Agile But Safe: Learning Collision-Free High-Speed Legged Locomotion},
 booktitle={Robotics: Science and Systems}, year={2024}}

@article{yang2022safety,
 author={Qisong Yang and Thiago D. Sim{\~a}o and Simon H. Tindemans and Matthijs T. J. Spaan},
 title={Safety-Constrained Reinforcement Learning with a Distributional Safety Critic},
 journal={Machine Learning}, volume={112}, number={3}, pages={859--887}, year={2023}}

@inproceedings{bellemare2017distributional,
 author={Marc G. Bellemare and Will Dabney and R{\'e}mi Munos},
 title={A Distributional Perspective on Reinforcement Learning},
 booktitle={International Conference on Machine Learning}, series={Proceedings of Machine Learning Research}, volume={70}, pages={449--458}, year={2017}}

@inproceedings{dabney2018distributional,
 author={Will Dabney and Mark Rowland and Marc G. Bellemare and R{\'e}mi Munos},
 title={Distributional Reinforcement Learning with Quantile Regression},
 booktitle={AAAI Conference on Artificial Intelligence}, year={2018}}

@inproceedings{dabney2018implicit,
 author={Will Dabney and Georg Ostrovski and David Silver and R{\'e}mi Munos},
 title={Implicit Quantile Networks for Distributional Reinforcement Learning},
 booktitle={International Conference on Machine Learning}, series={Proceedings of Machine Learning Research}, volume={80}, pages={1096--1105}, year={2018}}

@article{rockafellar2002conditional,
 author={R. Tyrrell Rockafellar and Stanislav Uryasev},
 title={Conditional Value-at-Risk for General Loss Distributions},
 journal={Journal of Banking \& Finance}, volume={26}, number={7}, pages={1443--1471}, year={2002}}

@inproceedings{schneider2024riskaware,
 author={Lukas Schneider and Jonas Frey and Takahiro Miki and Marco Hutter},
 title={Learning Risk-Aware Quadrupedal Locomotion Using Distributional Reinforcement Learning},
 booktitle={IEEE International Conference on Robotics and Automation}, year={2024}}

@inproceedings{shi2023robust,
 author={Jiyuan Shi and Chenjia Bai and Haoran He and Lei Han and Dong Wang and Bin Zhao and Mingguo Zhao and Xiu Li and Xuelong Li},
 title={Robust Quadrupedal Locomotion via Risk-Averse Policy Learning},
 booktitle={IEEE International Conference on Robotics and Automation}, year={2024}}

@inproceedings{chung2024add,
  title     = {Adversarial Environment Design via Regret-Guided Diffusion Models},
  author    = {Chung, Hojun and Lee, Junseo and Kim, Minsoo and Kim, Dohyeong and Oh, Songhwai},
  booktitle = {Advances in Neural Information Processing Systems},
  volume    = {37},
  pages     = {63715--63746},
  year      = {2024},
  doi       = {10.52202/079017-2035}
}

@article{mittal2025isaaclab,
  title   = {Isaac Lab: A GPU-Accelerated Simulation Framework for Multi-Modal Robot Learning},
  author  = {Mittal, Mayank and Roth, Pascal and Tigue, James and others},
  journal = {arXiv preprint arXiv:2511.04831},
  year    = {2025}
}

@article{schwarke2025rslrl,
  title   = {RSL-RL: A Learning Library for Robotics Research},
  author  = {Schwarke, Clemens and Mittal, Mayank and Rudin, Nikita
             and Hoeller, David and Hutter, Marco},
  journal = {arXiv preprint arXiv:2509.10771},
  year    = {2025}
}

@article{schulman2017ppo,
  title   = {Proximal Policy Optimization Algorithms},
  author  = {Schulman, John and Wolski, Filip and Dhariwal, Prafulla and Radford, Alec and Klimov, Oleg},
  journal = {arXiv preprint arXiv:1707.06347},
  year    = {2017}
}
